\documentclass{article}

\usepackage[preprint,nonatbib]{neurips_2024}

\usepackage[utf8]{inputenc} 
\usepackage[T1]{fontenc}    
\usepackage{amsfonts}       
\usepackage{nicefrac}       
\usepackage{xcolor}         
\usepackage{dsfont}
\usepackage{microtype}
\usepackage{hyperref}
\usepackage{url}
\usepackage{booktabs}
\usepackage[section]{placeins}

\usepackage[maxnames=5,natbib]{biblatex}
\usepackage{array}
\newcolumntype{L}[1]{>{\raggedright\let\newline\\\arraybackslash\hspace{0pt}}m{#1}}
\newcolumntype{C}[1]{>{\centering\let\newline\\\arraybackslash\hspace{0pt}}m{#1}}
\newcolumntype{R}[1]{>{\raggedleft\let\newline\\\arraybackslash\hspace{0pt}}m{#1}}

\AtEveryBibitem{%
\ifentrytype{article}{
    \clearfield{issn}
    \clearfield{url}%
    \clearfield{urlyear}%
    \clearfield{month}%
    \clearfield{pages}%
    \clearfield{volume}%
    \clearfield{number}%
}{}

\ifentrytype{inproceedings}{
    \clearfield{url}%
    \clearfield{urlyear}%
    \clearfield{month}
    \clearfield{address}%
    \clearfield{pages}%
    \clearfield{isbn}
}{}

\ifentrytype{misc}{
    \clearfield{primaryclass}%
    \clearfield{url}%
    \clearfield{issn}%
    \clearfield{urlyear}%
    \clearfield{month}
    \clearfield{journal}%
    \clearfield{doi}%
}{}
}

\DeclareSourcemap{
  \maps[datatype=bibtex]{
    \map{
      \step[fieldset=primaryclass,null]
    }
  }
}

\title{Do Large Language Models have Shared Weaknesses in Medical Question Answering?}

\author{%
Andrew M. Bean\thanks{Corresponding author: \{first\}.\{last\}@oii.ox.ac.uk} \quad Karolina Korgul \quad
Felix Krones \quad  Robert McCraith \quad Adam Mahdi \\\\
Oxford Internet Institute\\
University of Oxford\\
Oxford, United Kingdom\\
}

\begin{document}
\definecolor{darkblue}{rgb}{0, 0, 0.5}

\maketitle

\begin{abstract}
Large language models (LLMs) have made rapid improvement on medical benchmarks, but their unreliability remains a persistent challenge for safe real-world uses. To design for the use LLMs as a category, rather than for specific models, requires developing an understanding of shared strengths and weaknesses which appear across models. To address this challenge, we benchmark a range of top LLMs and identify consistent patterns across models. We test $16$ well-known LLMs on $874$ newly collected questions from Polish medical licensing exams. For each question, we score each model on the top-1 accuracy and the distribution of probabilities assigned. We then compare these results with factors such as question difficulty for humans, question length, and the scores of the other models. LLM accuracies were positively correlated pairwise ($0.39$ to $0.58$). Model performance was also correlated with human performance ($0.09$ to $0.13$), but negatively correlated to the difference between the question-level accuracy of top-scoring and bottom-scoring humans ($-0.09$ to $-0.14$). The top output probability and question length were positive and negative predictors of accuracy respectively (p~$< 0.05$). The top scoring LLM, GPT-4o Turbo, scored $84\%$, with Claude Opus, Gemini 1.5 Pro and Llama 3/3.1 between  $74$\% and $79\%$. We found evidence of similarities between models in which questions they answer correctly, as well as similarities with human test takers. Larger models typically performed better, but differences in training, architecture, and data were also highly impactful. Model accuracy was positively correlated with confidence, but negatively correlated with question length. We find similar results with older models, and argue that these patterns are likely to persist across future models using similar training methods.
\end{abstract}

\section{Introduction}

The healthcare field has historically been among the first places to test new methods in machine learning, with clear benefits for society when improvements can be made.
Large Language Models (LLMs) and foundation models broadly have attracted interest for their potential in medical applications  \citep{thirunavukarasuLargeLanguageModels2023}.
A rapidly increasing number of LLMs have been successful in medical question answering, achieving passing scores on benchmarks taken from medical licensing exams \citep{noriCapabilitiesGPT4Medical2023, tuGeneralistBiomedicalAI2023}.
However, LLMs still have fundamental issues with reliability, rooted in their nature as stochastic models trained on token prediction \citep{benderClimbingNLUMeaning2020, mccoyEmbersAutoregressionUnderstanding2023}, and in the unreliability of their training corpora, which is scraped from the internet and likely includes incorrect, outdated or deliberately misleading information \citep{duGLaMEfficientScaling2022, chowdheryPaLMScalingLanguage2022}.
These concerns, among others, make rigorous testing crucial for safe uses of LLMs. 

The first model to achieve a passing score\footnote{It is worth noting that the majority of humans taking these exams pass, and passing alone does not qualify someone as a doctor, so these scores cannot be equated with either human-level performance or medical proficiency.} on one of the standardised licensing exam datasets was Med-PaLM in 2022 \citep{tuGeneralistBiomedicalAI2023}.
Since then, other models have demonstrated the ability to score highly on these exams \citep{noriCapabilitiesGPT4Medical2023, singhalExpertLevelMedicalQuestion2023, tuGeneralistBiomedicalAI2023, tuConversationalDiagnosticAI2024}.
Specialist LLMs have also been created which are fine-tuned specifically for medical question answering and reach similar performance as the larger generalist models \citep{chenMEDITRON70BScalingMedical2023, christopheMed42ClinicalLarge2023, christopheMed42v2SuiteClinical2024}.
The large number of different LLMs being trained and released raises questions about how similar they are to each other, and the extent to which their reliability might be predictable \citep{brownLanguageModelsAre2020, openaiGPT4TechnicalReport2023, touvronLlamaOpenFoundation2023, chenMEDITRON70BScalingMedical2023, christopheMed42ClinicalLarge2023, dubeyLlamaHerdModels2024}.

New model releases typically include testing on well-known benchmarks to show that the model improves over existing options \citep{noriCapabilitiesGPT4Medical2023,chenMEDITRON70BScalingMedical2023,tuConversationalDiagnosticAI2024}.
This allows for comparisons between models in overall scores, but does not permit a comparison of the models on a per-question basis.
The most commonly used benchmarks, MedQA \citep{jinWhatDiseaseDoes2021}, MedMCQA \citep{palMedMCQALargescaleMultiSubject2022} and MMLU \citep{hendrycksMeasuringMassiveMultitask2021}, do include overall human performance metrics, but with a similar inability to compare on individual questions.
Our work enables comparisons both between LLMs and with humans in terms not only of \textit{how many} questions are answered correctly, but \textit{which ones}.
Doing so facilitates the identification of patterns in the types of questions which an LLM might be expected to answer correctly. 

In this paper, we conduct a comparison of 7 top LLMs and 9 older or smaller variants on a medical question answering dataset which uniquely allows us to compare LLM performance on a per question basis with medical students sitting the same exams.
This assessment allows us to identify patterns in LLMs' strengths and weaknesses which are consistent across different models, and also contrast those strengths and weaknesses with humans'.
These patterns contribute to the understanding of LLMs broadly, which is less driven by the particularities of any specific model.

\section{Methods}

\subsection{Dataset}
\paragraph{Format} The Medical Final Examination (Lekarski Egzamin Ko\'ncowy - LEK \citep{CentrumEgzaminowMedycznych}) is a Polish medical licensing examination, taken by doctors and other graduate medical trainees; it is available in both English and Polish \footnote{The questions are available at \url{https://cem.edu.pl/}}.
The examination is offered biannually in spring and autumn, and consists of 200 five-option multiple choice questions, where a passing score is 56\% correct, but average scores are around 80\%. For this study, we collected the questions used from the last five exam sittings, between spring 2021 and spring 2023.
After removing the 25 questions which were determined to be inaccurate by the examiners, and 101 repeated questions, the final dataset comprises 874 unique questions.

The questions cover ten categories, which we list with their counts: internal diseases (168), paediatrics (126), surgery (118), obstetrics and gynaecology (124), emergency medicine and intensive care (79), family medicine (87), psychiatry (61), bioethics and medical law (46), public health (34) and medical jurisprudence (31). This set of questions is similar in nature to other medical benchmarking datasets such as MedQA \citep{jinWhatDiseaseDoes2021}, and because the process of accessing the questions is highly manual,it is unlikely that the questions have been included in the training data of the LLMs being tested.

\paragraph{Human comparison} The key distinguishing factor of the LEK dataset is granular comparison to human results. About 8000 medical students and graduates have taken the exam each at each sitting for the past several years, and results are available for each individual question. This includes the percentage of test takers selecting each answer choice, an {\it index of difficulty} (IOD), which is the average accuracy of the top and bottom 27\% of test takers and a {\it discriminative power index} (DPI), which is the difference in accuracy between the top and bottom 27\%. 

\subsection{Models and implementation details}
\label{sec:implementation}

In Table~\ref{tab:models}, we describe each family of models tested to provide an overview of key existing models and highlight distinctive features of each. We include a complete listing of models tested and their parameters in Appendix~\ref{app:params}.

\begin{table}[t!]
\renewcommand{\arraystretch}{1.3} 
\centering
    \begin{tabular}{lL{.3\columnwidth}L{.4\columnwidth}}
        \hline
         \textbf{Provider} & \textbf{Models} & \textbf{Key Features} \\
         \hline
         OpenAI & GPT-4o \citep{openaiGPT4TechnicalReport2023}, GPT 3.5 \citep{ouyangTrainingLanguageModels2022} & Strong generalist models \\
         Google & Gemini 1.5 Pro \citep{geminiteamGeminiUnlockingMultimodal2024a} & A strong generalist model and the base model for multiple state-of-the-art medical models\\
         Anthropic & Claude Opus \citep{anthropicIntroducingNextGeneration} & A strong generalist model\\
         Mistral & Mixtral \citep{jiangMixtralExperts2024} & A unique architecture which combines smaller models to compete with larger models\\
         Meta & Llama~2 \citep{touvronLlamaOpenFoundation2023}, Llama 3 \citep{dubeyLlamaHerdModels2024}, Llama 3.1 & The base models for most open-access fine-tuned models \\
         M42 & Med42 v2 \citep{christopheMed42ClinicalLarge2023} & Fine-tuned from Llama~3 using medical texts\\
         EPFL & Meditron \citep{chenMEDITRON70BScalingMedical2023} & Fine-tuned from Llama~2 using medical texts \\
         Microsoft & Phi~3 \citep{abdinPhi3TechnicalReport2024} & A small model which uses high-quality data to compete with large models\\
         \hline
    \end{tabular}
    \smallskip
    \caption{\textbf{Included models.} The LLMs included in the study and details about their providers and motivation for inclusion.}
    \label{tab:models}
\end{table}

Models were accessed either via API or via Hugging Face and run locally. Hugging Face models were run on a server with 2 A100 GPUs and 2 L40 GPUs, with runtimes under 30 minutes per model. These models were quantised to optimise memory usage on the GPUs, which has a negligible performance impact \citep{frantarGPTQAccuratePostTraining2022}.

\subsection{Prompting}
Prompts for each question were generated by combining a base prompt: `{\it Answer the following multiple choice question by giving the most appropriate response.
The answer should be one of [A, B, C, D, E].}', with the content of the question and the multiple choice answers, followed by `{\it Answer:}'.
This structure is reflective of typical exam instructions.
For instruction-tuned models, we use any prompt templates recommended by the model provider. 
For closed models, we included an additional system prompt `\textit{You are a multiple choice answering chatbot. You only respond with single letters.}', in order to improve instruction-following. 
The answer choices were randomly reordered with a seed, so that any relationship between answer choice ordering and model performance is minimised, but kept consistent across models.
We use zero-shot prompting, which may impact the instruction-following abilities of the models, but should not impact the amount of encoded medical content\citep{kojimaLargeLanguageModels2022}.
While higher scores may be possible through further prompt optimisation \citep{yangLargeLanguageModels2023}, we believe that our approach is more representative, since most users are unlikely to carefully optimise their prompts.
Fully-formed prompts were submitted as a series of separate queries to each model.

\subsection{Evaluation metrics}\label{Sec:eval_metrics}

For open models, we used a bias term to condition the LLMs to only output one of the tokens corresponding to the five answer choices. For each of these LLMs, we extracted a probability distribution over the possible answers to each question [A,B,C,D,E], from which we select the highest probability response.

For closed models, we do not have the same control over the generations, and can only inspect the tokens being generated. For these models, we score accuracy based on whether or not the generated answer matches the correct answer. In cases where the model fails to follow the instructions and respond with a single letter, we mark the answer as incorrect.

\paragraph{Top-1 Accuracy} We calculated the {\it top-1 accuracy}, which we will also refer to as just {\it accuracy}, as the proportion of questions for which the language model's highest predicted probability matched the correct answer to the question. This is the answer which would be produced with greedy decoding (setting temperature to zero), and the typical metric for evaluating most multiple choice benchmarks \citep{bidermanLessonsTrenchesReproducible2024}.

\paragraph{Expected Accuracy} Since LLMs in general use do not typically use greedy decoding, we also consider the distribution of probabilities over the answer choices \citep{songGoodBadGreedy2024}. To capture this, we computed a per question expected accuracy score of the model, which is corresponds to the statistical notion of expected value. 
\[
{\rm Expected 
\,\,Accuracy}\,(model)\,\,\,= \sum_{c \in \{A,B,C,D,E\}} {\rm P}(c|model)\mathds{1}_{c^*}(c),
\]
where ${\rm P}(c|model)$ is the probability assigned to answer choice $c$ by a given $model$ and $\mathds{1}_{c^*}(c)$ is the binary indicator function of whether the answer $c$ was correct ($c^*$) or not. This simplifies to the probability that the LLM assigns to the correct answer. This metric is only available for the open models.

\subsection{Logistic regression models}

We fit logistic regressions to test for relationships between model accuracy and the predictive factors of question length and model confidence.
For each LLM we fit a logistic regression model
\[ \label{eq:logistic}
    {\rm P}(x) = 1/\big[1+e^{-(\beta_{0} + \beta_{1}x)}\big], 
\]
where ${\rm P}(x)$ is the probability that the specific LLM correctly answers a question with property $x$. We interpret $\beta_{1}$ as an indication of the direction of the relationship between the variables tested. For testing question length, we used the word count of the prompts as the predictor, since different models have different approaches to tokenisation. We then transformed the raw word count by taking the natural logarithm to account for the right skew of the distribution. For testing model confidence, we used the probability given to the highest likelihood response as the predictor, since this indicates whether model confidence is well-calibrated.

\section{Results}

In each subsection, we present complete results for the top models. We also describe the results for the older models tested, but the full results for older models are in Appendix~\ref{app:old_results}.

\subsection{Accuracy and expected accuracy scores}

\begin{table}[ht!]
\renewcommand{\arraystretch}{1.3} 
\centering
\renewcommand{\arraystretch}{1.3} 
\begin{tabular}
{lp{.08\columnwidth}p{.08\columnwidth}p{.08\columnwidth}p{.08\columnwidth}p{.08\columnwidth}p{.08\columnwidth}p{.08\columnwidth}p{.08\columnwidth}}
\hline
\textbf{Cat.} & \textbf{GPT 4o} & \textbf{Claude Opus} & \textbf{Gemini 1.5 Pro} & \textbf{Llama 3 70B} & \textbf{Llama 3.1 70B} & \textbf{Med 42 70B} & \textbf{Phi 3 Medium} & \textbf{Human} \\
\hline
BL & 73.9 & 67.4 & 69.6 & 58.7 & 52.2 & 60.9 & 52.2 & 	\textbf{83.6} \\
EM & \textbf{86.1} & 79.7 & 73.4 & 70.9 & 81.0 & 77.2 & 68.4 & 82.9 \\
FM & 74.7 & 73.6 & 64.4 & 69.0 & 64.4 & 60.9 & 50.6 & \textbf{81.6} \\
GS & \textbf{86.4} & 81.4 & 75.4 & 82.2 & 75.4 & 77.1 & 70.3 & 82.2 \\
ID & \textbf{87.5} & 79.2 & 76.8 & 77.4 & 75.0 & 73.2 & 67.9 & 83.3 \\
MJ & 80.6 & 74.2 & 64.5 & 58.1 & 67.7 & 64.5 & 61.3 & \textbf{84.1} \\
OB & \textbf{85.5} & 83.9 & 81.5 & 75.8 & 75.8 & 74.2 & 69.4 & 83.1 \\
PD & \textbf{82.5} & 77.0 & 77.8 & 71.4 & 74.6 & 72.2 & 72.2 & 81.5 \\
PH & \textbf{94.1} & 91.2 & 85.3 & 85.3 & 91.2 & 79.4 & 88.2 & 82.3 \\
PS & \textbf{86.9} & 83.6 & 82.0 & 83.6 & 82.0 & 75.4 & 75.4 & 75.6 \\
\hline
All & \textbf{84.2} & 79.3 & 75.7 & 74.6 & 74.3 & 72.3 & 67.6 & 82.1 \\
All (Ex.) & - & - & - & \textbf{73.6} & 63.7 & 72.1 & 66.1 & - \\
\hline
\end{tabular}
\smallskip
\caption{{\bf Human and LLM accuracies by question category.} The table compares human performance and seven LLMs across ten different medical categories (Cat.) of questions and in total (All). Bolded entries are the highest score in the category. Medical categories in the LEK database are (question count): internal disease ID (168), paediatrics PD (126), surgery GS (118), obstetrics and gynaecology OB (124), family medicine FM (87), psychiatry PS (61), emergency medicine and intensive care EM (79), bioethics and medical law BL (46), medical jurisprudence MJ (31) and public health PH (34). The final row of the table shows the expected accuracies for the open models, which are unavailable for the closed models.}
\label{tab:categories_main}
\end{table}

The accuracies of both the humans and LLMs across question categories from the LEK dataset are shown in Table~\ref{tab:categories_main}.
The models all exceeded the 56\% passing threshold on the exam, with GPT-4o slightly surpassing the average  performance of human test-takers.
The highest scoring model in each category was GPT-4o.
Across categories, most models had the weaker performance on the legal categories, medical jurisprudence and bioethics and law and the strongest performance on public health.

The Llama (and Med42) models scored similarly in accuracy to much larger closed source models, and Phi 3 came close to the Llama models despite only having 14B parameters.
For older models, we tested GPT 3.5, PaLM 2, Mixtral 8x7B, Llama 2 70B, and Meditron 70B, which had (top-1) accuracy scores $64.4$\%, $62.9$\%, $61.6$\%, $51.0$\% and $47.9$\%, respectively, worse than all of the more recent models of a similar size.
We also tested smaller versions of the open models and found that the smaller sized models of the same architecture were consistently weaker, though still above the passing mark. Llama~3 8B, Llama~3.1 8B, Med 42 8B, and Phi 3.5 Mini scored $60.1$\%, $57.6$\%, $62.8$\% and $59.5$\% respectively, similar to the larger older models.

The expected accuracies for the open models are shown in the last row of Table~\ref{tab:categories_main}. The models all had lower expected accuracy than top-1 accuracy, indicating that a non-deterministic generation would be less likely to be accurate than the top-1 score would indicate.

\begin{table}[ht!]
\centering
\renewcommand{\arraystretch}{1.2}
\begin{tabular}{lp{.08\columnwidth}p{.08\columnwidth}p{.08\columnwidth}p{.08\columnwidth}p{.08\columnwidth}p{.08\columnwidth}p{.08\columnwidth}p{.08\columnwidth}}
\hline
 & \textbf{GPT 4o} & \textbf{Claude Opus} & \textbf{Gemini 1.5 Pro} & \textbf{Llama 3 70B} & \textbf{Llama 3.1 70B} & \textbf{Med 42 70B} & \textbf{Phi 3 Medium} \\
\hline
\textbf{GPT 4o} & 1.00 & 0.58 & 0.45 & 0.48 & 0.47 & 0.45 & 0.40 \\
\textbf{Claude Opus} & 0.58 & 1.00 & 0.48 & 0.46 & 0.44 & 0.50 & 0.46 \\
\textbf{Gemini 1.5 Pro} & 0.45 & 0.48 & 1.00 & 0.45 & 0.42 & 0.50 & 0.48 \\
\textbf{Llama 3 70B} & 0.48 & 0.46 & 0.45 & 1.00 & 0.58 & 0.56 & 0.43 \\
\textbf{Llama 3.1 70B} & 0.47 & 0.44 & 0.42 & 0.58 & 1.00 & 0.47 & 0.39 \\
\textbf{Med 42 70B} & 0.45 & 0.50 & 0.50 & 0.56 & 0.47 & 1.00 & 0.47 \\
\textbf{Phi 3 Medium} & 0.40 & 0.46 & 0.48 & 0.43 & 0.39 & 0.47 & 1.00 \\
\hline
\end{tabular}
\smallskip
\caption{{\bf Question-level correlations.} Pairwise correlations on question-level top-1 accuracy between the main language models considered in this work. All correlations are significant with $p<0.05$.}
\label{tab:pairwise}
\end{table}

\subsection{Correlations with other LLMs}
\label{sec:model_corrs}
Table\,\,\ref{tab:pairwise} shows pairwise correlations on question-level top-1 accuracy between the models.
Every model was significantly correlated ($p<0.05$) to the other models, with correlations ranging from $0.39$ to $0.58$.
Llama~3 and Llama~3.1, the models with the most similar initial training and architecture, had the highest correlation, though GPT 4o and Claude Opus were similarly correlated.

For older models, correlations were positive but lower ($0.14$ to $0.46$). Correlations between the stronger old models, GPT-3.5, PaLM 2, and Mixtral 8x7B, and the current top models were in a similar range to the top models ($0.34$ to $0.46$).

\subsection{Correlations with human test takers}

As shown in Table~\ref{tab:difficulty_main}, every LLM showed a positive correlation with question difficulty (0.09 to 0.13) and average human scores (0.20 to 0.29).
The correlations with the discriminative power (DPI) were negative across tested language models (-0.09 to -0.14), indicating a tendency to score worse on questions which only top-scoring humans answered correctly.

In older models, the correlations to question difficulty were comparable for GPT-3.5, PaLM 2 and Mixtral 8x7B (0.12 to 0.14), but lower for Llama 2 70B and Meditron 70B (0.05). Correlations to human scores were in a slightly less positive (0.15 to 0.24), and correlation to the DPI was slightly more negative (-0.15 to -0.17).

\begin{table}[t!]
\centering
\renewcommand{\arraystretch}{1.2}
\begin{tabular}{{lp{.08\columnwidth}p{.08\columnwidth}p{.08\columnwidth}p{.08\columnwidth}p{.08\columnwidth}p{.08\columnwidth}p{.08\columnwidth}p{.08\columnwidth}}}
\hline
 & \textbf{GPT 4o} & \textbf{Claude Opus} & \textbf{Gemini 1.5 Pro} & \textbf{Llama 3 70B} & \textbf{Llama 3.1 70B} & \textbf{Med 42 70B} & \textbf{Phi 3 Medium} \\
\hline
IOD & 0.12 & 0.12 & 0.12 & 0.11 & 0.12 & 0.13 & 0.09 \\
DI & -0.12 & -0.13 & -0.09 & -0.14 & -0.11 & -0.13 & -0.13 \\
Human & 0.23 & 0.22 & 0.21 & 0.20 & 0.20 & 0.24 & 0.29 \\
\hline
\smallskip
\end{tabular}
\caption{\textbf{Correlation of LLM accuracy and human question performance.} The first two rows show correlations between the expected accuracy scores of the LLMs and the two human difficulty metrics. The third row shows the correlation with the average human scores on each question. The index of difficulty (IOD) is the average score of the top and bottom 27\% of human examinees, while the discriminative power index (DPI) is the difference in the average scores between the top and bottom 27\% of human examinees. All correlations are significant with $p<0.05$.}
\label{tab:difficulty_main}
\end{table}

\subsection{Predictive factors}
As shown in Table \ref{tab:length}, GPT-4o, and Claude Opus did not have significant relationships  between accuracy and question length. The other models had negative relationships, particularly Phi 3. Older models also had negative relationships.

Every open model had a positive relationship between the probability assigned by the model to the top response and the accuracy of the model, indicating that the models were more likely to be correct when the response was given a higher probability. This was also the case for the older models tested.

\begin{table}[ht!]
\centering
\small
\renewcommand{\arraystretch}{1.3} 
\begin{tabular}{lp{.08\columnwidth}p{.08\columnwidth}p{.08\columnwidth}p{.08\columnwidth}p{.08\columnwidth}p{.08\columnwidth}p{.08\columnwidth}p{.08\columnwidth}}
\hline
 & \textbf{GPT 4o} & \textbf{Claude Opus} & \textbf{Gemini 1.5 Pro} & \textbf{Llama 3 70B} & \textbf{Llama 3.1 70B} & \textbf{Med 42 70B} & \textbf{Phi 3 Medium} \\
\hline
${\rm ln}(words)$ & -0.28  & -0.20  & -0.46* & -0.46* & -0.48* & -0.41* & -0.71* \\
$\max({\rm P}(token))$ & -  & -  & - & 1.59* & 1.03* & 1.92* & 1.32* \\
\hline
\multicolumn{2}{l}{\footnotesize{* p $< 0.05$}}
\end{tabular}
\caption{\textbf{Logistic regression coefficients for predictive variables.} For each model, we fit a logistic regression between the correctness of the answer for each question (0 or 1) and the natural logarithm of the number of words in the question, and a regression between the correctness of the answer and the probability assigned to the top response. The values shown here correspond to $\beta_1$ in equation~\ref{eq:logistic}.}
\label{tab:length}
\end{table}

\section{Discussion}

\subsection{Main finding}

We found that larger LLMs outperformed smaller LLMs of their own family, but that differences between model families can be a more important determinant of model performance than size.
For most models, we found that performance was related to the length of the question and the probability the model placed on its response, though the top models had less decrease in performance on longer questions.
All of the models were positively correlated with the other models and with human accuracy on the same question.
Med 42, the specially fine-tuned medical model performed similarly to the other models, but did not improve the Llama 3 model it was trained from, and top generalist models were better than the specialist model as well.
We observe that these patterns are also present in previous generations of models, sometimes to a lesser degree.
Due to shared LLM architectures, training data and loss functions, we anticipate that many future models may also follow similar patterns.

\subsection{Possible implications for health data science}
\label{sec:implications}
From these results, we identify four patterns which appear across LLMs.
We then connect to mechanisms from existing research and properties of the models to show why these patterns might persist with future models. Each of these patterns exists in both current and prior generations of models, and across model sizes.

\medskip
\noindent
{\it Larger models may have higher accuracy, but data and architecture definitely matter}

As in previous studies \citep{kaplanScalingLawsNeural2020, hoffmannTrainingComputeOptimalLarge2022}, with the LEK dataset, GPT-4o, thought to be the largest model, achieved the highest accuracy scores (Table~\ref{tab:categories_main}).
Within model families, Llama~3/3.1 70B, Med 42 70B, and Phi 3 Medium also substantially outperformed smaller versions of the same models.
However, Gemini 1.5 Pro, Llama 3/3.1 and Phi 3 all scored similarly despite being different sizes.
Phi 3 exemplifies the potential for high-quality training data to enable smaller models to be competitive consistent with similar results for other tasks \citep{mitraOrcaTeachingSmall2023}.

\noindent
{\it LLMs have shared strengths and weaknesses}

Across models, we found weak evidence of shared strengths and weaknesses (Table~\ref{tab:pairwise}).
Previous works have offered reasons for this including similar subject coverage in the training data \citep{carliniExtractingTrainingData2021, solaimanProcessAdaptingLanguage2021, leeDeduplicatingTrainingData2022}, linguistic patterns within questions \citep{mccoyRightWrongReasons2019} or statistical artefacts such as answer choice ordering \citep{pezeshkpourLargeLanguageModels2023}.
The correlations between Med 42, Llama 3 and Llama~3.1, which differ the least, were similar to the correlations between GPT 4o and Claude Opus, models with much less in common. 

\medskip
\noindent
{\it LLMs are better at questions that humans find easier}

Across LLMs, performance was better on questions which humans found easier and worse on questions which only the top humans were able to answer (Table~\ref{tab:difficulty_main}).
This could indicate a similarity between what makes questions difficult for humans and LLMs, potentially due to factors such as the training being based on human feedback data and human text \citep{mccoyEmbersAutoregressionUnderstanding2023}.
To the extent that humans and LLMs have similar expertise, the marginal value of the LLM as an assitant is reduced, since it does not contribute new information or ability \citep{wilderLearningComplementHumans2020}.

Across models, medical jurisprudence was a particularly weak category.
This may be explainable by the locality of the data, as training data will likely include jurisprudence from countries around the world, and particularly from the United States, while the questions focus only on the Polish context, where laws may differ.
Such local particularities will be a challenge for the training and deployment of LLMs in a global context, as training data cannot be easily associated with a single country of origin or relevance.

\medskip
\noindent
{\it{Model accuracy is partially predictable}} 

The logistic regressions both showed consistent patterns across models, with increasing prompt length corresponding  to lower scores, and higher top probabilities corresponding to higher scores (Table~\ref{tab:length}).
Previous work has shown that longer contexts can also lead to poor processing of the input information \citep{liuLostMiddleHow2023}.
This effect was less pronounced for the top models, (GPT-4o, Claude Opus), indicating that the improvements in these models are partially in responding to longer questions.

Tests of whether language models are well-calibrated, predicting higher probabilities for more likely responses, have had mixed results in previous studies \citep{kadavathLanguageModelsMostly2022, jiangHowCanWe2021}. 
The results in this study did show a directional relationship, but most models consistently put most of the probability on a single answer, which does not translate directly into a confidence measure or probability of being correct.
As such, while we find evidence of some degree of calibration, the particulars are more varied than for other patterns identified.

\subsection{Limitations}
\label{sec:limitations}
For our analysis, we have focused on accuracy on multiple choice questions as a measure of model performance.
This approach forces the LLMs to produce one of five valid responses, of which one is certain to be correct and may therefore avoid errors such as omitting relevant details in open-ended responses.
The particular choice of prompt can also impact how well LLMs perform a task \citep{yangLargeLanguageModels2023, sclarQuantifyingLanguageModels2023}.
We found the results to be consistent when tested with an alternate prompt, but undiscovered prompts may exist which have greater variation.

Within the scope of this paper, we also faced practical limitations which reduced the applicability of results.
Although Gemini 1.5 Pro is included, the specialist Med-Gemini model is not available for public testing \citep{saabCapabilitiesGeminiModels2024a}.
We have sought to capture a wide range of LLMs which apply a variety of different approaches in order to increase the applicability of the results, but it remains possible that new methods will depart from existing paradigms.

We have specifically focused on comparisons between LLMs in their ability to answer medical questions.
This is not a complete picture of the possible issues of relevance to a medical practitioners for the adoption of LLMs in medicine, which may also include possible conceptual challenges to the use of language models \citep{benderClimbingNLUMeaning2020} as well as  bias in outputs and error patterns \citep{parrishBBQHandBuiltBias2022}.


\printbibliography

\appendix

\section{Results for Additional Models}
\label{app:old_results}

We include the complete results for the previous generation models referenced in the main body of the paper here. For these models, the closed models did support access to the probabilities for the top outputs, so the expected accuracy scores are included for all models.

\begin{table}[ht!]
\centering
\renewcommand{\arraystretch}{1.2} 
\begin{tabular}{lccccc}
\hline
 \textbf{Cat.} & \textbf{GPT 3.5} & \textbf{PaLM 2} & \textbf{Llama 2 70B} & \textbf{Meditron 70B} & \textbf{Mixtral 8x7B} \\
\hline
BL & 54.3 & 54.3 & 54.3 & 41.3 & 	\textbf{63.0} \\
EM & 65.8 & 57.0 & 48.1 & 45.6 & \textbf{68.4} \\
FM & 55.2 & \textbf{56.3} & 49.4 & 44.8 & 52.9 \\
GS & 65.3 & \textbf{69.5} & 52.5 & 46.6 & 64.4 \\
ID & \textbf{66.1} & 59.5 & 47.0 & 52.4 & 57.7 \\
MJ & 41.9 & \textbf{48.4} & 41.9 & 35.5 & \textbf{48.4} \\
OB & \textbf{67.7} & 64.5 & 54.0 & 50.8 & 61.3 \\
PD & 65.1 & \textbf{65.9} & 48.4 & 43.7 & 60.3 \\
PH & 82.4 & \textbf{85.3} & 79.4 & 61.8 & 79.4 \\
PS & \textbf{70.5} & 68.9 & 50.8 & 52.5 & 68.9 \\
\hline
All & \textbf{64.4} & 62.9 & 51.0 & 47.9 & 61.6 \\
All (Ex.) & \textbf{63.7} & 53.9 & 50.9 & 36.4 & 61.5 \\
\hline
\end{tabular}
\smallskip
\caption{{\bf Human and LLM accuracies by question category for additional models.} Comparison of five LLMs across ten different medical categories (Cat.) Medical categories in the LEK database are (question count): internal disease ID (168), paediatrics PD (126), surgery GS (118), obstetrics and gynaecology OB (124), family medicine FM (87), psychiatry PS (61), emergency medicine and intensive care EM (79), bioethics and medical law BL (46), medical jurisprudence MJ (31) and public health PH (34). The final row of the table shows the expected accuracies for the open models.}
\end{table}

\begin{table}[ht!]
    \centering
    \renewcommand{\arraystretch}{1.2}
\begin{tabular}{llllll}
\hline
 & \textbf{GPT 3.5} & \textbf{PaLM 2} & \textbf{Llama 2 70B} & \textbf{Meditron 70B} & \textbf{Mixtral 8x7B} \\
 \hline
GPT 3.5 & 1.00* & 0.42* & 0.35* & 0.24* & 0.38* \\
GPT 4o & 0.33* & 0.37* & 0.29* & 0.22* & 0.36* \\
Claude Opus & 0.35* & 0.41* & 0.31* & 0.19* & 0.43* \\
PaLM 2 & 0.42* & 1.00* & 0.40* & 0.33* & 0.41* \\
Gemini 1.5 Pro & 0.37* & 0.34* & 0.23* & 0.15* & 0.40* \\
Meditron 70B & 0.24* & 0.33* & 0.40* & 1.00* & 0.21* \\
Llama 2 70B & 0.35* & 0.40* & 1.00* & 0.40* & 0.31* \\
Llama 3 8B & 0.42* & 0.45* & 0.35* & 0.32* & 0.39* \\
Llama 3 70B & 0.36* & 0.34* & 0.23* & 0.22* & 0.34* \\
Llama 3.1 8B & 0.44* & 0.45* & 0.26* & 0.31* & 0.39* \\
Llama 3.1 70B & 0.37* & 0.37* & 0.18* & 0.21* & 0.33* \\
Med 42 8B & 0.41* & 0.43* & 0.30* & 0.23* & 0.43* \\
Med 42 70B & 0.39* & 0.40* & 0.32* & 0.22* & 0.45* \\
Mixtral 8x7B & 0.38* & 0.41* & 0.31* & 0.21* & 1.00* \\
Phi 3 Medium & 0.46* & 0.46* & 0.34* & 0.23* & 0.46* \\
Phi 3.5 Mini & 0.44* & 0.42* & 0.40* & 0.26* & 0.38* \\
\hline
\end{tabular}
\smallskip
\caption{{\bf Pairwise correlations on question-level top-1 accuracy for additional models.}}
\end{table}

\begin{table}[ht!]
    \centering
    \renewcommand{\arraystretch}{1.2}
\begin{tabular}{llllll}
\hline
 & \textbf{GPT 3.5} & \textbf{PaLM 2} & \textbf{Llama 2 70B} & \textbf{Meditron 70B} & \textbf{Mixtral 8x7B} \\
 \hline
IOD & 0.14* & 0.12* & 0.05  & 0.05  & 0.12* \\
DI & -0.16* & -0.13* & -0.16* & -0.17* & -0.15* \\
Human & 0.24* & 0.19* & 0.18* & 0.15* & 0.21* \\
\hline
\end{tabular}
\smallskip
\caption{\textbf{Correlation of LLM accuracy and human question performance for additional models.}}
\end{table}

\begin{table}[ht]
\centering
\renewcommand{\arraystretch}{1.2}
\begin{tabular}{llllll}
\hline
 & \textbf{GPT 3.5} & \textbf{PaLM 2} & \textbf{Llama 2 70B} & \textbf{Meditron 70B} & \textbf{Mixtral 8x7B} \\
 \hline
ln(words) & -0.57* & -0.37* & -0.13  & -0.16  & -0.39* \\
$\max({\rm P}(token))$ & 1.35* & 0.77* & 1.08* & 0.66* & 1.20* \\
\hline
\end{tabular}
\smallskip
\caption{\textbf{Logistic regression coefficients for predictive variables for additional models.}}
\end{table}

\section{Model Parameters}
\label{app:params}

Table~\ref{tab:hyperparams} shows the hyperparameters used for each of the models tested in this study. Full code is available on \hyperlink{https://github.com/am-bean/llm-med-comparisons}{GitHub}.
\begin{table}[ht!]
    \centering
    \renewcommand{\arraystretch}{1.2}
    \begin{tabular}{llccc}
    \hline
          & \textbf{Version} & \textbf{Temp.} & \textbf{Max Tokens} & \textbf{Quantization} \\
         \hline
         GPT-4o & gpt-4o-2024-05-13 & 0 & 1 & - \\
         GPT-3.5 & gpt-3.5-turbo-1106 & 0 & 1 & - \\
         Gemini 1.5 Pro & gemini-1.5-pro & 0 & 1 & - \\
         PaLM 2 & text-bison@002 & 0 & 1 & - \\
         Claude 3 Opus & claude-3-opus-20240229 & 0 & 2 & - \\
         Llama 3 70B & meta-llama/Meta-Llama-3-70B-Instruct & 0 & 1 & 4 bit\\
         Llama 3 8B & meta-llama/Meta-Llama-3-8B-Instruct & 0 & 1 & 8 bit\\
         Llama 3.1 70B & meta-llama/Meta-Llama-3.1-70B-Instruct & 0 & 1 & 4 bit\\
         Llama 3.1 8B & meta-llama/Meta-Llama-3.1-8B-Instruct & 0 & 1 & 8 bit\\
         Llama 2 70B & meta-llama/Llama-2-70B-chat-hf & 0 & 1 & 4 bit\\
         Med42 v2 70B & m42-health/Llama3-Med42-70B & 0 & 1 & 4 bit\\
         Med42 v2 8B & m42-health/Llama3-Med42-8B & 0 & 1 & 8 bit\\
         Meditron 70B & epfl-llm/meditron-70b & 0 & 1 & 4 bit\\
         Phi 3 Medium & microsoft/Phi-3-medium-4k-instruct & 0 & 1 & 8 bit\\
         Phi 3.5 Mini & microsoft/Phi-3.5-mini-instruct & 0 & 1 & None \\
         Mixtral 8x7B & mistralai/Mixtral-8x7B-Instruct-v0.1 & 0 & 1 & 4 bit\\
         \hline
    \end{tabular}
    \smallskip
    \caption{\textbf{Model hyperparameters.}}
    \label{tab:hyperparams}
\end{table}



\end{document}